\pdfoutput=1

\documentclass[11pt]{article}

\usepackage[preprint]{acl}

\usepackage{times}
\usepackage{latexsym}

\usepackage[T1]{fontenc}

\usepackage[utf8]{inputenc}

\usepackage{microtype}

\usepackage{inconsolata}

\usepackage{graphicx}

\usepackage{tabularray}
\usepackage{subfigure}
\usepackage[normalem]{ulem}

\usepackage[most]{tcolorbox}
\tcbuselibrary{listingsutf8}

\usepackage{linguex}

\usepackage{longtable}
\usepackage{booktabs}

\title{Steering Prepositional Phrases in Language Models: A Case of\\\textit{with}-headed Adjectival and Adverbial Complements in \texttt{Gemma}-2}

\author{Stefan Arnold \and Rene Gröbner \\ 
Friedrich-Alexander-Universität Erlangen-Nürnberg \\ Lange Gasse 20, 90403 Nürnberg, Germany \\ 
\texttt{(stefan.st.arnold, rene.edgar.gröbner)@fau.de}}

\begin{document}
\maketitle
\begin{abstract}

Language Models, when generating prepositional phrases, must often decide for whether their complements functions as an instrumental adjunct (describing the verb adverbially) or an attributive modifier (enriching the noun adjectivally), yet the internal mechanisms that resolve this split decision remain poorly understood. In this study, we conduct a targeted investigation into \texttt{Gemma}-2 to uncover and control the generation of prepositional complements. We assemble a prompt suite containing \textit{with}-headed prepositional phrases whose contexts equally accommodate either an instrumental or attributive continuation, revealing a strong preference for an instrumental reading at a ratio of $3$:$4$. To pinpoint individual attention heads that favor instrumental over attributive complements, we project activations into the vocabulary space. By scaling the value vector of a single attention head, we can shift the distribution of functional roles of complements, attenuating instruments to 33\% while elevating attributes to 36\%.

\end{abstract}

\section{Introduction}
\label{sec:introduction}

Transformer-based \citep{vaswani2017attention} Language Models (LMs) \citep{devlin2019bert, brown2020language, chowdhery2023palm} internalize rich inventories of dependency grammar \citep{jawahar2019bert, hu2020systematic} and deploy this structural knowledge to generate grammatically coherent sentences. Targeted evaluations showed that these models reliably choose the correct lexeme from a set of grammatically minimally different continuations in variety of syntactic constructions, including \textit{agreement} \citep{linzen2016assessing, goldberg2019assessing, finlayson2021causal}, \textit{licensing} \citep{wilcox018rnn, warstadt2019investigating}, \textit{binding} \citep{marvin2018targeted}, and the structure of \textit{arguments} \citep{kann2019verb, conia2022probing}.

\begin{figure}[!tb]
    \centering
    \includegraphics[width=0.45\textwidth]{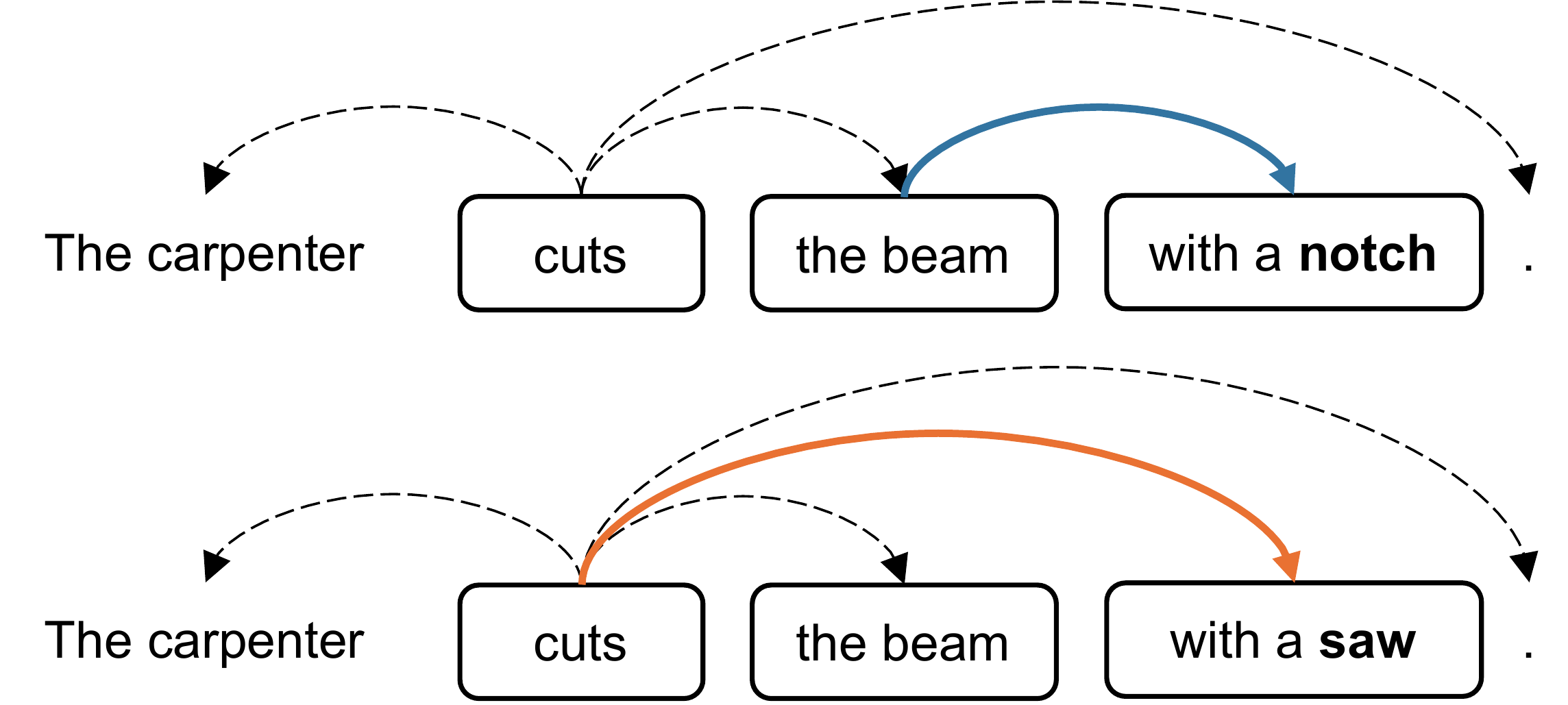}
    \caption{Example for the ambiguity of prepositional phrases: $\uparrow$ attaches the complement (\textit{notch}) to the noun phrase (\textit{beam}), acting as a adjectival modifier that supplies an attribute, whereas $\downarrow$ attaches the complement (\textit{saw}) to the verb phrase (\textit{cut}), serving as an adverbial modifier that introduces an instrument.
    }
    \label{fig:template}
\end{figure}

In this study, we examine modifier attachment in prepositional phrases (PP). A PP typically contains a preposition as head immediately followed by a complement, which can serve as either an adverbial modifier or adjectival modifier \citep{nakashole2015knowledge}. Figure \ref{fig:template} presents an illustrative example of a PP with distinct functional roles of their complements. In the adverbial role, the PP (i.e., \textit{saw}) attaches to the verb phrase, denoting an instrument that describes an action. In the adjectival role, the PP (i.e., \textit{notch}) attaches to the noun phrase, providing an attribute to an object. Both functional roles can compete during generation, and in some scenarios it might be desirable that the model prioritizes an instrumental reading, while in other scenarios, it might be more desirable to emphasize an attributive reading. Yet the internal mechanisms that govern whether an autoregressive LM produces a verb-modifying or noun-modifying adjunct have not been probed.

Recent work in the field of mechanistic interpretability \citep{wang2023interpretability, geiger2025causal} is aimed at reverse engineering the internal processes carried out by LMs in terms of intelligible circuits. These circuits represent faithful simplifications of opaque computations, allowing us to interpret the contributions of individual model components to the final generated text. Circuit-level analyses have recently pinpointed components responsible for performing induction \citep{olsson2022context, wang2023interpretability} and reasoning \citep{brinkmann2024mechanistic}, storing and retrieving factual knowledge \citep{meng2022editing, geva2023dissecting, merullo2023characterizing, wang2025functional}, and enforcing structural well-formedness \citep{finlayson2021causal}. 



\paragraph{Contribution.} To dissect a circuit that characterizes the selectional preferences for preposition‑led modifiers, we make three key contributions.

\begin{enumerate}

\item We design a controlled task by manually authoring prompts centered on \textit{with}‑introduced PP, whose context evokes both an instrument and an attribute, forcing a model to choose between these equally plausible but competing completions. We then assess this task on \texttt{Gemma}-2 \citep{riviere2024gemma} and note a preference toward adverbial instruments over adjectival attributes, occurring in a ratio of $3$:$4$.

\item To localize the promotion of instrumentive and attributive continuations, we employ logit attribution \citep{belrose2023eliciting}. By projecting internal activations into the vocabulary space, this technique allows us to make claims about the model components most responsible for favoring instruments or attributes. When applied to attention heads, we reveal a single head that consistently delivers the dominant direction towards instruments.

\item Through activation scaling \citep{merullo2023characterizing}, we demonstrate a targeted intervention that reliably steers prepositional complements. By applying a scalar to the value of the attention head, we can shift the biased preference into a near-balanced distribution of instrumentive and attributive modifiers. We are able to raise the rate of attributes to 36\% while reducing the amount of instruments to 33\% by downweighting a single attention head, adjusting only a minimal number of parameters.

\end{enumerate}

\section{Related Work}


\subsection{Model Introspection}

The interpretability community has devised a diverse set of techniques for understanding internal representations. These techniques span a spectrum of evidence: from \textit{behavioral evaluations} that infer internals from output observations, through \textit{structural correlations} that relate internals to formal linguistics, to \textit{causal interventions} that manipulate internals to edit model behavior.

\paragraph{Behavioral evaluations} study model outputs under carefully designed stimuli to reveal syntactic rules or semantic abilities. \citet{linzen2016assessing} assembled texts containing curated stimuli for which perplexity is evaluated as evidence of the presence or absence of linguistic knowledge.

\paragraph{Structural interpretations} aim to identify linguistic properties captured in hidden states through auxiliary models applied at sentence-level \citep{adi2016fine, conneau2018you} and word-level \citep{tenney2019you}, while \citet{hewitt2019structural} particularly designed a structural probe to recover parse trees from hidden states. 

Our work is built upon probing via vocabulary projections \citep{ghandeharioun2024patchscopes} in which hidden representations are inspected in the vocabulary space by mapping them directly through the unembedding matrix rather than an auxiliary model. By applying the identity function, we can map representations from any layer to the final layer, interpreting every hidden state into a distribution over the vocabulary. Extensions apply linear \citep{pal2023future} or affine \citep{belrose2023eliciting} mapping to improve the interpretability in the vocabulary space. 

\paragraph{Causal interventions} seek to locate the mechanisms that mediate behavior. By contrasting clean and corrupted inputs and patching intermediate activations from the clean run into the corrupted run, \textit{activation patching} \citep{vig2020investigating, meng2022editing} puts forth an intervention for isolating functional circuits. The granularity of localization spans individual neurons \citep{finlayson2021causal}, attention heads \citep{merullo2023characterizing, wang2023interpretability}, feedforward sublayers \citep{meng2022editing}, and the residual stream \citep{belrose2023eliciting}.


Apart from localization, interventions can also be used for model steering. \citet{meng2023massediting} modify weight matrices of feedforward layers to edit factual associations, whereas \citet{rimsky2024steering} add a learned direction to the attention heads.


Recent interpretability efforts have derived functional mechanism responsible for performing tasks described in-context \citep{olsson2022context, hou2023towards, brinkmann2024mechanistic, singh2024needs}, storage and recall of factual knowledge and other forms of memories \citep{geva2023dissecting, merullo2023characterizing, wang2025functional}, and concrete mechanism tailored to linguistic inflection \citep{finlayson2021causal}, arithmetic calculation \citep{ stolfo2023mechanistic}, and numerical comparison \citep{hanna2023does}. These findings reinforce that transformer internals can be decomposed into reusable motifs: attention heads copy and move information \cite{wang2023interpretability}, feedforward layers serve as key–value memories \citep{geva2021transformer}, and the residual stream linearly accumulates the contributions from all model components. 

\subsection{Phrase Attachment}

PP is typically framed from a parsing perspective, concerned with deciding whether a PP modifies the verb phrase or the noun phrase. Ambiguity arises because both attachments are often syntactically valid when semantic priors that license any reading are absent \citep{karamolegkou2025trick}.

To disambiguate PP attachment, approaches rely on lexicalization \citep{ratnaparkhi1994maximum, pantel2000unsupervised}, contextualization \citep{ratnaparkhi2021resolving}, and the integration of world knowledge \citep{nakashole2015knowledge}. Although PP attachment has been studied extensively in parsing, the mechanisms by which LMs express selectional preferences over PP complements have not undergone a targeted analysis. We recast \textbf{PP attachment} to \textbf{PP completion} where the model must choose between two unambiguous adjuncts, both of which are made plausible by the preceding context.



\begin{table*}[!tbh]
\centering
\resizebox{\linewidth}{!}{%
\begin{tblr}{
  cell{1}{1} = {c=3}{l}, 
  hline{1,7} = {-}{0.08em},
  hline{2} = {1-3}{0.03em},
  hline{2} = {4-5}{lr},
}
\textbf{Prompts with licensing contexts}                 &                                 &                                                      & \textbf{Instrument} & \textbf{Attribute} \\
A carpenter has a \textit{saw}.  & A beam has a \textit{notch}.    & The carpenter \uline{cuts} the \uline{beam with} a   & \textbf{saw}                 & notch              \\
A chef has a \textit{syringe}.   & A cake has a \textit{frosting}. & The chef \uline{decorated} the \uline{cake with} a   & \textbf{syringe}             & frosting           \\
A florist has a \textit{shear}.  & A bouquet has a \textit{rose}.  & The florist \uline{trims} the \uline{bouquet with} a & \textbf{shear}              & rose               \\
A pilot has a \textit{joystick}. & A plane has a \textit{failure}. & The pilot \uline{lands} the \uline{plane with} a     & \textbf{joystick}            & failure         \\
A welder has a \textit{torch}.   & A joint has a \textit{crack}.   & The welder \uline{seals} the \uline{joint with} a    & \textbf{torch}               & crack              
\end{tblr}
}
\caption{Excerpt from our prompt suite. \textit{italic} indicates the candidate instrument and attribute; \underline{underline} represents the phrase triplet containing a verb, noun, and preposition for which the model must decide whether it attaches the instrument or attribute; \textbf{bold} highlights the preferred prepositional complement.}
\label{tab:prompts}
\end{table*}

\section{Task Formulation}

PP attachment is a classic problem in syntactic parsing, concerned with deciding whether a PP modifies the verb phrase (high attachment) or the noun phrase (low attachment). The ambiguity arises because both options are often syntactically licit in the absence of licensing context \citep{karamolegkou2025trick}. Given a PP occurring within a sentence where multiple attachment sites are possible, the goal is to select the most plausible site.

We formalize PP attachment through an ordered quadruple \((V,N,P,C)\), where \(V\) is the main verb, \(N\) is the head noun of its direct object, \(P\) is the preposition, and \(C\) is the candidate complement of the PP. The task is to decide whether the PP \((P,C)\) attaches \emph{adverbially} to \(V\) (instrumental adjunct) or \emph{attributively} to \(N\) (nominal modifier). These roles yield distinct structures and meanings: an \emph{instrument} specifies how the action is performed, whereas an \emph{attribute} describes a property of the noun. Prior work typically treats the PP as a unit to be attached over a closed set of candidate parses.

To study the mechanisms LMs employ when context licenses contrastive PP continuations, we recast PP attachment as a generative zero-shot decision. Rather than choosing among candidate parses, the model must \emph{continue} a \textit{with}-headed PP with either an instrument or an attribute. This framing serves as a representative probe of selectional preferences. Our experimental design relies on stimuli in which role attribution of the PP adjunct is unequivocal. Because PP corpora feature equivocal constructions, we needed to construct a manually curated set of prompts in which each continuation is unambiguous in its syntactic role, yet the surrounding context is crafted to render both options equally coherent and contextually plausible.





We reformulate PP attachment for continuation so that the model is tasked with producing the PP complement. To make both readings viable while keeping the options functionally distinct, we manually curate pairs of complements that are \emph{each} unambiguous in role identity but jointly create a context in which either completion is plausible. We constrain prompts to a structured subject-verb-object frame followed by a \textit{with}-headed PP:


\ex. The \textit{subject} \textit{verb} the \textit{object} with a \dots

This placement ensures that both instrumental and attributive continuations are plausible and the design naturally suits zero-shot prompting in which the model generates the PP complement directly.

Because PP complement generation depends on world knowledge, we augment the prompts with a minimal licensing context that associates the subject with a plausible instrument and the object with a plausible attribute. Each context introduces two entities, where the subject-associated noun acts as plausible \emph{instrument} for the action and the object-associated noun as  plausible \emph{attribute} of the object. 

\ex. A \textit{subject} has a \textit{subject-associated noun}.

\ex. A \textit{object} has a \textit{object-associated noun}.

For our experiments, we select \texttt{Gemma}-2, an autoregressive model with two billion parameters. \texttt{Gemma}-2 is an enticing candidate for probing mechanisms of selectional preference due to its large vocabulary size with numerous reserved words, and studying it also contributes to the growing body of interpretation for the \texttt{Gemma} family \citep{lieberum2024gemma}.

Table \ref{tab:prompts} presents selected examples of our manually licensed prompts. Appendix \ref{appendix} provides the complete set of prompts, comprising a total of 100. We observe that the \texttt{Gemma}-2 model manifests instrumental adjuncts rather than attributive modifier. This observation is consistent with psycholinguistic studies reporting that humans tend to favor high attachment to verbs over low attachment to nouns \citep{spivey1995resolving}, and with recent findings indicating that language models display a bias toward instrumental rather than attributive readings \citep{zhou2024tree}.


\section{Logit Attribution}

We aim to isolate a faithful mechanism within our language model that governs the selectional preferences for the formation of PP.

To pinpoint model components driving this selectional preference for an adverbial or adjectival reading, we turn to \textit{logit attribution} \citep{belrose2023eliciting}. The idea behind logit attribution is to interpret the role of a particular component in a language model for a given task in terms of the vocabulary space. This is built on the premise that the residual stream can be decomposed into the sum of contributions from every model component. Recall that each model component in the transformer adds its output onto the residual stream, and the residual stream state gets projected onto the unembedding matrix, producing the logits distribution. Due to the linearity of the residual stream, every layer of computation can be traced back as the direct effect of each sublayer to the logits up to that point.

Because PP continuation in our task boils down to copying one of two lexical adjuncts from a context that licenses an instrument and attribute equally plausibly, we apply logit attribution to attention heads. This choice is motivated by the key role of attention heads in performing copying operations \citep{wang2023interpretability, merullo2023characterizing}.

We obtain the direct effects of attention heads favoring instrumental or attributive readings following \citet{merullo2023characterizing}. We start by extracting the corresponding vectors in the unembedding matrix for the target adjuncts. The additive update made by the attention layer is composed of the concatenated updates of each attention head after it is passed through the output weight matrix within the attention layer. We therefore divide the weight matrix of the attention output into one component for each attention head and project the head activations into the space of the residual stream by multiplying them with the corresponding slice of weight matrix. We then dot product the projected activation of the attention head with the weight vectors of the attribute and instrument from the unembedding matrix, giving us a scalar value representing the logit for each of those continuations represented by the head. We then compute the dot product between the projected activation of a given head and the unembedding vector for each target word, yielding a scalar logit for each adjunct. By subtracting these two logit values, we get the direct attribution to the logit difference between attribute and instrument. This logit difference captures the effect the head has in promoting one word (relative to another) to be the continuation: a positive value indicates that the head writes in the direction of the attribute, promoting an adjectival reading, whereas a negative value indicates that the head writes in the direction of the instrument, promoting an adverbial reading.

\begin{figure}[!tbh]
    \centering
    \includegraphics[width=0.45\textwidth]{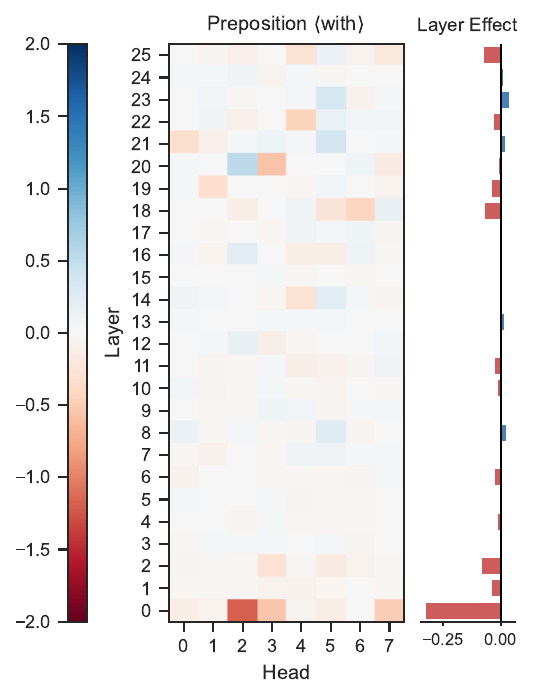}
    \caption{Attribution map of direct effects for \textit{with}-headed PP constructions. Only a single attention head shows a strong selectional preference.}
    \label{fig:ppa_mha}
\end{figure}

Figure \ref{fig:ppa_mha} visualizes the logit difference calculation for each head in every layer. Since \texttt{Gemma}-2 has 26 layers and 8 heads for each layer, this totals 208 heads to test. Despite some variation in the roles of every head throughout our prompt suite, we identify a series of heads that push the model towards attributive modifiers or instrumental adjuncts. However, the heads that consistently affect PP completion are clustered in early layers, and these heads uniformly drive the model toward instrumental adjuncts. \texttt{L0H2} emerges as the principal driver of \textit{with}-headed phrase completions, rendering it an ideal target for steering interventions.


\begin{figure}[!tbh]
    \centering
    \includegraphics[width=0.45\textwidth]{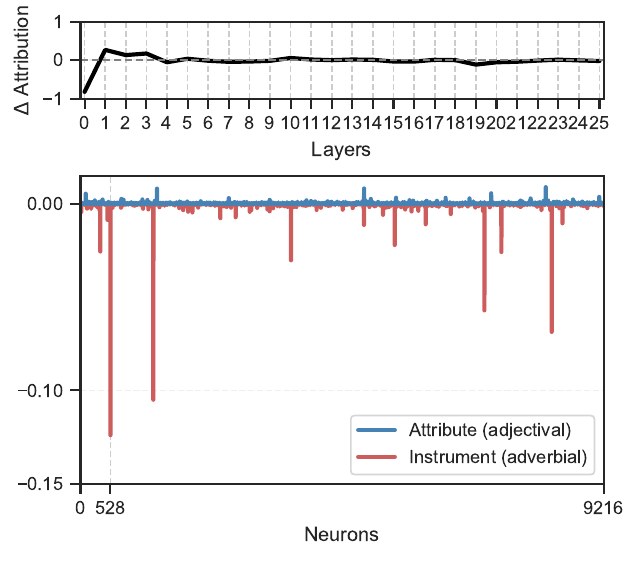} 
    \caption{Attribution across feedforward layers. Only a sparse subset of neurons contribute to the choice between attributive and instrumental readings.}
    \label{fig:ppa_mlp}
\end{figure}

To validate that attention heads provide the primary signals for PP continuation, we extended our attribution analysis to feedforward layers. Figure \ref{fig:ppa_mlp} plots the averaged preference for all 26 layers along with its scores for the 9216 neurons (only for the initial layer). We observe that most feedforward layers show no consistent directional bias, contributing similarly to both readings. Any non‐zero attributions are almost exclusively confined to initial feedforward layer, where only a handful of neurons register meaningful effects on the instrumental versus attributive difference in logits, and even these contributions are an order of magnitude smaller than those we observed for attention heads. This lack of directional specificity in neuron attributions allows us to exclude feedforward sublayers as primary drivers of PP completion decisions.

\section{Activation Steering}

Several techniques have been proposed to steer the generative process of LMs. \citet{merullo2023characterizing} scale the activation of an attention head by a scalar, whereas \citet{rimsky2024steering} add a learned direction to the activation of an attention head. Since a single attention head heavily contributes to the direction in logit, we adopt the scaling intervention. 

We hypothesize that downweighting \texttt{L0H2} will enable us to suppress instrumental readings and boost attributive readings. To test our hypothesis, we apply a multiplicative factor $\alpha < 0$ to the value vector of \texttt{L0H2}. The effect of this intervention is measured by the proportion of times the model flips the functional role of the PP complement from an instrumental modifier to an attributive modifier. 

\begin{figure}[!tb]
    \centering
    \includegraphics[width=0.45\textwidth]{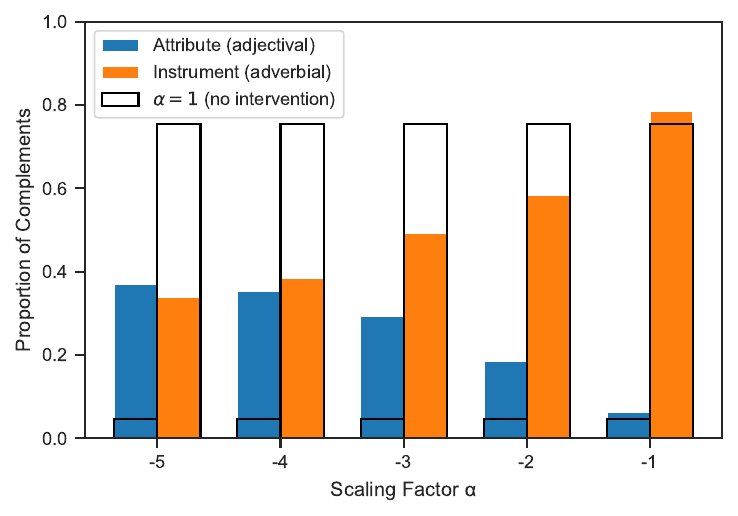}
    \caption{Proportion of shifts in selectional preferences by the multiplicative value under activation scaling.}
    \label{fig:intervention}
\end{figure}

Figure \ref{fig:intervention} presents the proportions of instruments and attributes as a function of the scaling factor  $\alpha \in [-5,-4,-3,-2,-1]$. Without any intervention, the model selects the instrument in 75\%, an attribute in 4\%, and deploys unlicensed words as PP complements in 21\%, suggesting a marked bias in the functional roles of PP complements. We find that scaling down attention head \texttt{L0H2} has a strong effect on flipping prepositional complements \footnote{We note that scaling other attention heads (not depicted in Figure \ref{fig:intervention}) produces markedly weaker shifts, underscoring the unique role of \texttt{L0H2} on selectional preferences.}. We can attenuate instruments to 33\% while elevating attributes to 36\%, as we decrease the multiplicative value of $\alpha$. However, even extreme downweighting does not eliminate instrumental readings entirely, and a substantial portion of completions fall into alternative adjuncts outside the two intended options, indicating that aggressive steering can divert the model toward unlicensed complements.

\section{Conclusion}

Through a controlled prompt design, logit attribution, and activation scaling, we isolated and manipulated a single attention head in \texttt{Gemma}-2 that exerts a dominant influence on the balance between adverbial and adjectival continuations following from prepositional phrases. Our findings provide a principled proof of concept for steering model output in contexts where multiple continuations are syntactically and semantically plausible, bridging interpretable and controllable generation. This study advances our understanding of how autoregressive models internally resolve functional role preferences and demonstrates that such steering can be achieved with minimal parameter interventions.


\paragraph{Limitations.} We acknowledge two main limitations. First, we conduct all experiments on a single model. This choice is motivated by its large vocabulary size, which facilitates controlled and targeted mechanistic analysis. However, it does not guarantee that our identified mechanism generalizes to other model architectures. Second, we restrict our investigation to PPs as a representative case of selectional preference under ambiguity. Although this scope offers a controlled and well-documented testing ground, it represents only one narrow subset of the model’s broader selectional preferences. 

We plan to extend our mechanistic understanding of selectional preferences of ambiguous continuations to garden-path effects~\citep{amouyal2025lm} and control over the assignment of predicate-argument structure like \textit{purpose}, \textit{location}, or \textit{time}, going beyond adverbial and adjectival modifiers.


\bibliography{custom}

\clearpage
\onecolumn
\appendix
\section{Prompt Suite} 
\label{appendix}
\begingroup
\footnotesize
\setlength{\tabcolsep}{3pt} 
\renewcommand{\arraystretch}{1.25}

\begin{longtable}{p{0.24\textwidth} p{0.22\textwidth} p{0.30\textwidth} p{0.24\textwidth}}
\toprule
\textbf{Subject-Instrument} & \textbf{Object-Attribute} & \textbf{Subject-Verb-Object} & \textbf{Adjuncts} \\
\midrule
\endfirsthead
\toprule
\textbf{Subject-Instrument} & \textbf{Object-Attribute} & \textbf{Subject-Verb-Object} & \textbf{Adjuncts} \\
\midrule
\endhead
\bottomrule \multicolumn{4}{r}{\small\itshape Table continued.} \\
\endfoot
\bottomrule
\endlastfoot
⟨baker, whisk⟩ & ⟨bowl, lump⟩ & ⟨baker, stirs, bowl⟩ & ⟨whisk, lump⟩ \\
⟨banker, spreadsheet⟩ & ⟨portfolio, stock⟩ & ⟨banker, edits, portfolio⟩ & ⟨spreadsheet, stock⟩ \\
⟨barber, scissor⟩ & ⟨beard, fringe⟩ & ⟨barber, trims, beard⟩ & ⟨scissor, fringe⟩ \\
⟨barista, portafilter⟩ & ⟨cappuccino, foam⟩ & ⟨barista, prepares, cappuccino⟩ & ⟨portafilter, foam⟩ \\
⟨bartender, shaker⟩ & ⟨cocktail, garnish⟩ & ⟨bartender, prepares, cocktail⟩ & ⟨shaker, garnish⟩ \\
⟨biologist, pipette⟩ & ⟨tube, liquid⟩ & ⟨biologist, tranfers, tube⟩ & ⟨pipette, liquid⟩ \\
⟨bonesetter, splint⟩ & ⟨patient, fracture⟩ & ⟨bonesetter, stabilizes, patient⟩ & ⟨splint, fracture⟩ \\
⟨brewer, keg⟩ & ⟨beer, trademark⟩ & ⟨brewer, dispenses, beer⟩ & ⟨keg, trademark⟩ \\
⟨builder, spatula⟩ & ⟨wall, crack⟩ & ⟨builder, repairs, wall⟩ & ⟨spatula, crack⟩ \\
⟨butcher, cleaver⟩ & ⟨steak, marbling⟩ & ⟨butcher, cuts, steak⟩ & ⟨cleaver, marbling⟩ \\
⟨carpenter, saw⟩ & ⟨beam, notch⟩ & ⟨carpenter, cuts, beam⟩ & ⟨saw, notch⟩ \\
⟨carpenter, chisel⟩ & ⟨plank, groove⟩ & ⟨carpenter, deepens, plank⟩ & ⟨chisel, groove⟩ \\
⟨cartographer, compass⟩ & ⟨map, legend⟩ & ⟨cartographer, aligns, map⟩ & ⟨compass, legend⟩ \\
⟨chef, ladle⟩ & ⟨pot, broth⟩ & ⟨chef, serves, pot⟩ & ⟨ladle, broth⟩ \\
⟨chef, spoon⟩ & ⟨egg, shell⟩ & ⟨chef, cracks, egg⟩ & ⟨spoon, shell⟩ \\
⟨chef, syringe⟩ & ⟨cake, frosting⟩ & ⟨chef, decorates, cake⟩ & ⟨syringe, frosting⟩ \\
⟨chef, spatula⟩ & ⟨meal, marinade⟩ & ⟨chef, flips, meal⟩ & ⟨spatula, marinade⟩ \\
⟨chef, spice⟩ & ⟨soup, flavor⟩ & ⟨chef, seasons, soup⟩ & ⟨spice, flavor⟩ \\
⟨chemist, pipette⟩ & ⟨reaction, precipitate⟩ & ⟨chemist, measures, reaction⟩ & ⟨pipette, precipitate⟩ \\
⟨chemist, flask⟩ & ⟨reaction, catalyst⟩ & ⟨chemist, conducts, reaction⟩ & ⟨flask, catalyst⟩ \\
⟨cleaner, vacuum⟩ & ⟨carpet, crumb⟩ & ⟨cleaner, cleans, carpet⟩ & ⟨vacuum, crumb⟩ \\
⟨coach, whistle⟩ & ⟨team, streak⟩ & ⟨coach, signals, team⟩ & ⟨whistle, streak⟩ \\
⟨conductor, baton⟩ & ⟨orchestra, listener⟩ & ⟨conductor, directs, orchestra⟩ & ⟨baton, listener⟩ \\
⟨cosmologist, telescope⟩ & ⟨planet, moon⟩ & ⟨cosmologist, observes, planet⟩ & ⟨telescope, moon⟩ \\
⟨cosmonaut, spacesuit⟩ & ⟨capsule, porthole⟩ & ⟨cosmonaut, abandones, capsule⟩ & ⟨spacesuit, porthole⟩ \\
⟨dentist, mirror⟩ & ⟨tooth, cavity⟩ & ⟨dentist, examines, tooth⟩ & ⟨mirror, cavity⟩ \\
⟨designer, tablet⟩ & ⟨product, stamp⟩ & ⟨designer, creates, product⟩ & ⟨tablet, stamp⟩ \\
⟨detective, lens⟩ & ⟨scene, clue⟩ & ⟨detective, inspects, scene⟩ & ⟨lens, clue⟩ \\
⟨diver, camera⟩ & ⟨reef, fish⟩ & ⟨diver, captures, reef⟩ & ⟨camera, fish⟩ \\
⟨doctor, thermometer⟩ & ⟨child, disease⟩ & ⟨doctor, checks, child⟩ & ⟨thermometer, disease⟩ \\
⟨draughtsman, ruler⟩ & ⟨blueprints, balcony⟩ & ⟨draughtsman, edits, blueprints⟩ & ⟨ruler, balcony⟩ \\
⟨driver, wheel⟩ & ⟨road, curve⟩ & ⟨driver, navigates, road⟩ & ⟨wheel, curve⟩ \\
⟨driver, wrench⟩ & ⟨car, tire⟩ & ⟨driver, repairs, car⟩ & ⟨wrench, tire⟩ \\
⟨farmer, plow⟩ & ⟨field, furrow⟩ & ⟨farmer, cuts, field⟩ & ⟨plow, furrow⟩ \\
⟨firefighter, ladder⟩ & ⟨cat, collar⟩ & ⟨firefighter, saves, cat⟩ & ⟨ladder, collar⟩ \\
⟨fisherman, net⟩ & ⟨crab, shell⟩ & ⟨fisherman, captures, crab⟩ & ⟨net, shell⟩ \\
⟨florist, shear⟩ & ⟨bouquet, rose⟩ & ⟨florist, trims, bouquet⟩ & ⟨shear, rose⟩ \\
⟨gardener, rake⟩ & ⟨garden, tree⟩ & ⟨gardener, grooms, garden⟩ & ⟨rake, tree⟩ \\
⟨gardener, can⟩ & ⟨plant, stem⟩ & ⟨gardener, waters, plant⟩ & ⟨can, stem⟩ \\
⟨gardener, shovel⟩ & ⟨soil, worms⟩ & ⟨gardener, digs, soil⟩ & ⟨shovel, worms⟩ \\
⟨gardener, shears⟩ & ⟨hedge, nest⟩ & ⟨gardener, prunes, hedge⟩ & ⟨shears, nest⟩ \\
⟨gardener, spade⟩ & ⟨garden, border⟩ & ⟨gardener, outlines, garden⟩ & ⟨spade, border⟩ \\
⟨geologist, scale⟩ & ⟨rock, fissure⟩ & ⟨geologist, measures, rock⟩ & ⟨scale, fissure⟩ \\
⟨guard, weapon⟩ & ⟨property, fence⟩ & ⟨guard, protects, property⟩ & ⟨weapon, fence⟩ \\
⟨hunter, rifle⟩ & ⟨forest, deer⟩ & ⟨hunter, targets, forest⟩ & ⟨rifle, deer⟩ \\
⟨janitor, mop⟩ & ⟨floor, scuffing⟩ & ⟨janitor, cleans, floor⟩ & ⟨mop, scuffing⟩ \\
⟨jeweler, cloth⟩ & ⟨ring, diamond⟩ & ⟨jeweler, examines, ring⟩ & ⟨cloth, diamond⟩ \\
⟨journalist, recorder⟩ & ⟨politician, controversy⟩ & ⟨journalist, interviews, politician⟩ & ⟨recorder, controversy⟩ \\
⟨judge, hammer⟩ & ⟨trial, verdict⟩ & ⟨judge, concludes, trial⟩ & ⟨hammer, verdict⟩ \\
⟨laboratorian, centrifuge⟩ & ⟨sample, contamination⟩ & ⟨laboratorian, separates, sample⟩ & ⟨centrifuge, contamination⟩ \\
⟨lawyer, highlighter⟩ & ⟨contract, clause⟩ & ⟨lawyer, reviews, contract⟩ & ⟨highlighter, clause⟩ \\
⟨librarian, scanner⟩ & ⟨book, cover⟩ & ⟨librarian, catalogs, book⟩ & ⟨scanner, cover⟩ \\
⟨lifeguard, whistle⟩ & ⟨swimmer, monofin⟩ & ⟨lifeguard, signals, swimmer⟩ & ⟨whistle, monofin⟩ \\
⟨lineman, multimeter⟩ & ⟨circuit, voltage⟩ & ⟨lineman, tests, circuit⟩ & ⟨multimeter, voltage⟩ \\
⟨locksmith, dietrich⟩ & ⟨keyway, vulnerability⟩ & ⟨locksmith, enters, keyway⟩ & ⟨dietrich, vulnerability⟩ \\
⟨magician, wand⟩ & ⟨mirage, misdirection⟩ & ⟨magician, conjures, mirage⟩ & ⟨wand, misdirection⟩ \\
⟨mason, level⟩ & ⟨wall, cladding⟩ & ⟨mason, trues, wall⟩ & ⟨level, cladding⟩ \\
⟨mathematician, chalkboard⟩ & ⟨formula, mistake⟩ & ⟨mathematician, derives, formula⟩ & ⟨chalkboard, mistake⟩ \\
⟨mechanic, wrench⟩ & ⟨engine, leak⟩ & ⟨mechanic, fixes, engine⟩ & ⟨wrench, leak⟩ \\
⟨midwife, doppler⟩ & ⟨woman, complication⟩ & ⟨midwife, monitors, woman⟩ & ⟨doppler, complication⟩ \\
⟨miner, axe⟩ & ⟨rock, gem⟩ & ⟨miner, breaks, rock⟩ & ⟨axe, gem⟩ \\
⟨musician, tuner⟩ & ⟨piece, pitch⟩ & ⟨musician, tunes, piece⟩ & ⟨tuner, pitch⟩ \\
⟨musician, turntable⟩ & ⟨song, rhythm⟩ & ⟨musician, streches, song⟩ & ⟨turntable, rhythm⟩ \\
⟨neurologist, penlight⟩ & ⟨pupil, dilation⟩ & ⟨neurologist, assesses, pupil⟩ & ⟨penlight, dilation⟩ \\
⟨nurse, syringe⟩ & ⟨arm, vaccine⟩ & ⟨nurse, treats, arm⟩ & ⟨syringe, vaccine⟩ \\
⟨painter, brush⟩ & ⟨wall, patch⟩ & ⟨painter, overpaints, wall⟩ & ⟨brush, patch⟩ \\
⟨painter, roller⟩ & ⟨canvas, sketch⟩ & ⟨painter, brushes, canvas⟩ & ⟨roller, sketch⟩ \\
⟨performer, script⟩ & ⟨scene, prop⟩ & ⟨performer, enters, scene⟩ & ⟨script, prop⟩ \\
⟨pharmacist, mortar⟩ & ⟨leaf, stem⟩ & ⟨pharmacist, grinds, leaf⟩ & ⟨mortar, stem⟩ \\
⟨photographer, flash⟩ & ⟨portrait, shadow⟩ & ⟨photographer, illuminates, portrait⟩ & ⟨flash, shadow⟩ \\
⟨photographer, camera⟩ & ⟨scene, horizon⟩ & ⟨photographer, captures, scene⟩ & ⟨camera, horizon⟩ \\
⟨physiotherapist, spirometer⟩ & ⟨patient, symptom⟩ & ⟨physiotherapist, screens, patient⟩ & ⟨spirometer, symptom⟩ \\
⟨pilot, joystick⟩ & ⟨plane, failure⟩ & ⟨pilot, controls, plane⟩ & ⟨joystick, failure⟩ \\
⟨plumber, wrench⟩ & ⟨pipe, leak⟩ & ⟨plumber, seals, pipe⟩ & ⟨wrench, leak⟩ \\
⟨policeman, handcuff⟩ & ⟨criminal, scar⟩ & ⟨policeman, arrests, criminal⟩ & ⟨handcuff, scar⟩ \\
⟨prehistorian, shovel⟩ & ⟨fossil, patina⟩ & ⟨prehistorian, excavates, fossil⟩ & ⟨shovel, patina⟩ \\
⟨programmer, keyboard⟩ & ⟨database, password⟩ & ⟨programmer, accesses, database⟩ & ⟨keyboard, password⟩ \\
⟨programmer, debugger⟩ & ⟨codebase, bug⟩ & ⟨programmer, debugs, codebase⟩ & ⟨debugger, bug⟩ \\
⟨ranger, tranquilizer⟩ & ⟨tiger, wound⟩ & ⟨ranger, paralyzes, tiger⟩ & ⟨tranquilizer, wound⟩ \\
⟨receptionist, telephone⟩ & ⟨visitor, question⟩ & ⟨receptionist, calls, visitor⟩ & ⟨telephone, question⟩ \\
⟨roofer, harness⟩ & ⟨rope, knot⟩ & ⟨roofer, fastens, rope⟩ & ⟨harness, knot⟩ \\
⟨scientist, microscope⟩ & ⟨slide, specimen⟩ & ⟨scientist, examines, slide⟩ & ⟨microscope, specimen⟩ \\
⟨sculptor, chisel⟩ & ⟨block, grain⟩ & ⟨sculptor, carves, block⟩ & ⟨chisel, grain⟩ \\
⟨singer, microphone⟩ & ⟨stage, spotlight⟩ & ⟨singer, performs, stage⟩ & ⟨microphone, spotlight⟩ \\
⟨sniper, scope⟩ & ⟨hideout, threat⟩ & ⟨sniper, targets, hideout⟩ & ⟨scope, threat⟩ \\
⟨statistician, notebook⟩ & ⟨datasets, schema⟩ & ⟨statistician, analyzes, datasets⟩ & ⟨notebook, schema⟩ \\
⟨stenographer, headset⟩ & ⟨speech, message⟩ & ⟨stenographer, transcribes, speech⟩ & ⟨headset, message⟩ \\
⟨student, pen⟩ & ⟨textbook, diagram⟩ & ⟨student, marks, textbook⟩ & ⟨pen, diagram⟩ \\
⟨surgeon, knife⟩ & ⟨tumor, mass⟩ & ⟨surgeon, removes, tumor⟩ & ⟨knife, mass⟩ \\
⟨tailor, needle⟩ & ⟨suit, tear⟩ & ⟨tailor, mends, suit⟩ & ⟨needle, tear⟩ \\
⟨tailor, thread⟩ & ⟨fabric, seam⟩ & ⟨tailor, stitches, fabric⟩ & ⟨thread, seam⟩ \\
⟨tailor, tape⟩ & ⟨dress, pattern⟩ & ⟨tailor, measures, dress⟩ & ⟨tape, pattern⟩ \\
⟨teacher, pointer⟩ & ⟨presentation, figure⟩ & ⟨teacher, points, presentation⟩ & ⟨pointer, figure⟩ \\
⟨teacher, chalk⟩ & ⟨board, equation⟩ & ⟨teacher, writes, board⟩ & ⟨chalk, equation⟩ \\
⟨topographer, theodolite⟩ & ⟨bridge, camber⟩ & ⟨topographer, surveys, bridge⟩ & ⟨theodolite, camber⟩ \\
⟨translator, dictionary⟩ & ⟨text, term⟩ & ⟨translator, translates, text⟩ & ⟨dictionary, term⟩ \\
⟨vet, stethoscope⟩ & ⟨pet, stroke⟩ & ⟨vet, monitors, pet⟩ & ⟨stethoscope, stroke⟩ \\
⟨waiter, cloth⟩ & ⟨table, decoration⟩ & ⟨waiter, cleans, table⟩ & ⟨cloth, decoration⟩ \\
⟨welder, torch⟩ & ⟨joint, crack⟩ & ⟨welder, seals, joint⟩ & ⟨torch, crack⟩ \\
⟨writer, pen⟩ & ⟨manuscript, flaw⟩ & ⟨writer, corrects, manuscript⟩ & ⟨pen, flaw⟩ \\
\end{longtable}
\endgroup

\end{document}